\documentclass{article}

\usepackage{PRIMEarxiv}

\usepackage[utf8]{inputenc} 
\usepackage[T1]{fontenc}    
\usepackage{hyperref}       
\usepackage{url}            
\usepackage{booktabs}       
\usepackage{amsfonts}       
\usepackage{nicefrac}       
\usepackage{microtype}      
\usepackage{lipsum}
\usepackage{natbib}
\usepackage{graphicx}
\usepackage{amsmath}
\usepackage[table]{xcolor}
\usepackage{tabularx}
\usepackage{siunitx}
\usepackage{colortbl}
\usepackage{makecell}
\usepackage{threeparttable}
\usepackage{multirow}
\usepackage{adjustbox} 
\usepackage{array}
\usepackage{graphicx}
\usepackage{caption}
\usepackage{subcaption}
\usepackage{comment}
\usepackage{fontawesome5}
\definecolor{ModelGreen}{RGB}{213,232,212}
\newcolumntype{L}[1]{>{\raggedright\arraybackslash}p{#1}}

\title{\Large Fleming-VL: Towards Universal Medical Visual Reasoning with Multimodal LLMs
}

\author{
Yan Shu \quad Chi Liu \quad Robin Chen  \quad Derek Li \quad Bryan Dai\thanks{Corresponding author} \\
Ubiquant \\
\texttt{\{shuyan9812,mzchen2001\}@gmail.com} \\
\texttt{\{cliu04,jdli,cbdai\}@ubiquant.com}
}

\begin{document}
\maketitle

\vspace{-1.0cm}


\vspace*{0.5cm}

\begin{abstract}
Multimodal Large Language Models (MLLMs) have demonstrated remarkable effectiveness in various general-domain scenarios, such as visual question answering and image captioning. Recently, researchers have increasingly focused on empowering MLLMs with medical conversational abilities, which hold significant promise for clinical applications. However, medical data presents unique challenges due to its heterogeneous nature—encompassing diverse modalities including 2D images (e.g., X-rays, CT slices, ultrasound, dermoscopy), 3D volumetric scans (e.g., CT, MRI), and temporal video sequences (e.g., surgical videos, echocardiography). The substantial domain gap and data format inconsistencies across these modalities have hindered the development of unified medical MLLMs. To address these challenges, we propose Fleming-VL, a unified end-to-end framework for comprehensive medical visual understanding across heterogeneous modalities. Fleming-VL tackles this problem from a data-centric perspective through three key strategies. First, we scale up the pretraining process by integrating long-context data from both natural images and medical-specific domains to establish robust foundational capabilities. Second, during fine-tuning, we strategically complement the training with rare medical data, including holistic video analysis and underrepresented 2D modalities such as ultrasound and dermoscopy images. Third, we extend existing medical multimodal evaluation frameworks to incorporate 3D volumetric and video understanding benchmarks. Through supervised fine-tuning (SFT) and group relative policy optimization (GRPO), we develop Fleming-VL in multiple model scales. Extensive experiments demonstrate that Fleming-VL achieves state-of-the-art performance across multiple benchmarks, including medical VQA, video QA, and 3D medical image understanding. We publicly release Fleming-VL to promote transparent, reproducible, and auditable progress in medical AI, enabling safer deployment in high-stakes clinical environments.

\end{abstract}

\begin{figure}[h]
    \centering
    \includegraphics[width=0.98\textwidth]{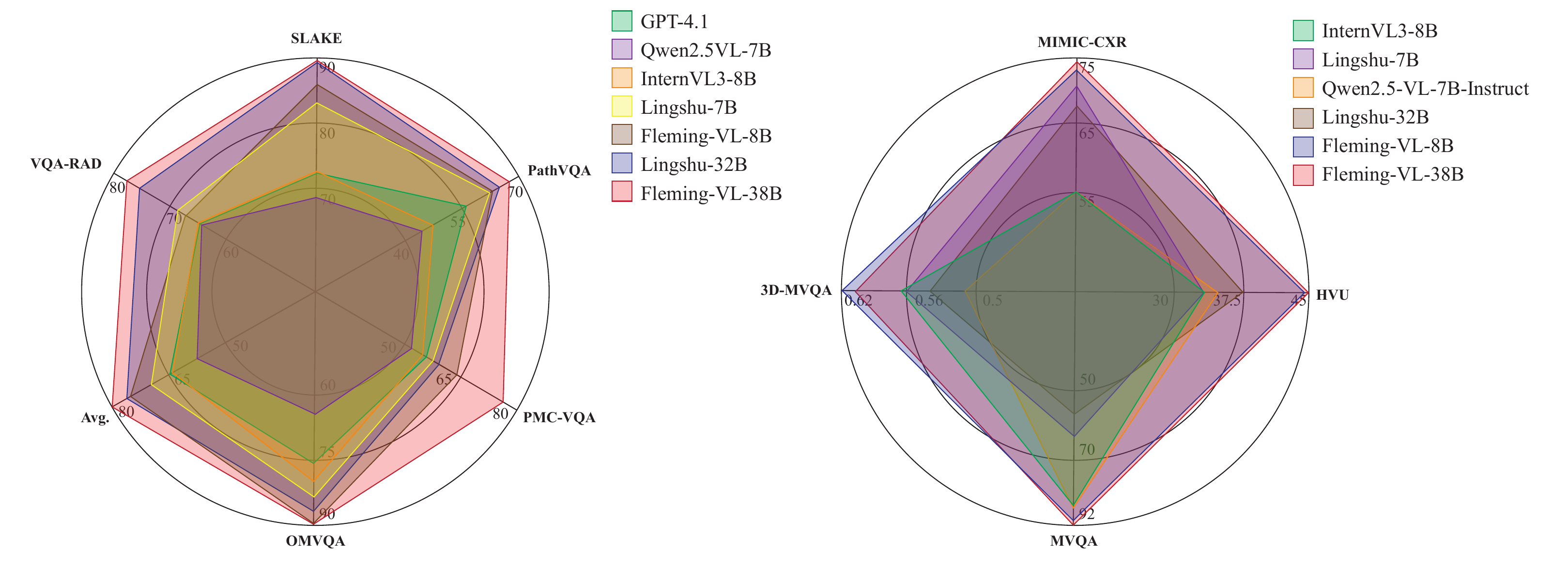}
    \caption{Medical multimodal understanding benchmarks performance. ``HVU'', ``MVQA'' and ``3D-MVQA'' denote holistic video understanding, medical video question answering and 3D-medical video question answering.}
    \label{fig:scatter-1}
\end{figure}


\section{Introduction}
Multimodal Large Language Models (MLLMs) \cite{liu2023visual,liu2024improved,yao2024minicpm}  have emerged as a transformative paradigm in artificial intelligence, seamlessly integrating visual and linguistic understanding through a unified architecture. The prevailing approach involves employing vision encoders to extract visual representations, which are then aligned with pre-trained Large Language Models (LLMs) via projection layers or adapter modules, enabling the model to process and reason over multimodal inputs in an end-to-end manner. This architectural design has demonstrated remarkable success across various general-domain tasks, including visual question answering (VQA), image captioning, and visual reasoning, with models such as LLaVA \cite{liu2023visual}, InstructBLIP \cite{dai2023instructblip}, and Qwen-VL \cite{wang2024qwen2} achieving human-level performance on numerous benchmarks. Building upon these successes, the research community has increasingly directed efforts toward adapting MLLMs for specialized medical applications, recognizing their potential to revolutionize healthcare delivery. Recent works \cite{li2023llava,li2024llava,xu2025lingshu,medgemini2024,liu2025fleming,wang2023huatuo} have explored diverse medical scenarios, including automated medical report generation from radiological images, disease diagnosis and classification, clinical decision support, and patient-doctor conversational assistance. These adaptations typically involve continued pre-training on medical image-text pairs and fine-tuning on task-specific medical datasets, aiming to bridge the domain gap between natural and medical images while preserving the strong reasoning capabilities inherited from general-domain MLLMs.

Despite these promising advances, current medical MLLMs face significant challenges that limit their practical utility in real-world clinical settings. A fundamental limitation is the inability to simultaneously understand diverse medical visual modalities within a unified framework. Medical data inherently spans multiple dimensions, including 2D images (such as X-rays, pathology slides, ultrasound, and dermoscopy), 3D volumetric scans \cite{bai2024m3d,chen20243d} (such as CT and MRI), and temporal video sequences \cite{chen2025surgllm} (such as surgical recordings and echocardiography). Existing approaches typically specialize in a single modality, requiring separate models for different data types, which not only increases deployment complexity but also prevents the model from leveraging cross-modal knowledge transfer. Furthermore, even within the same modality category, severe data imbalance poses a critical challenge. Medical MLLMs exhibit strong performance on data-rich modalities like chest X-rays and CT scans, but suffer from substantial performance degradation on underrepresented modalities such as ultrasound imaging and dermoscopy. This data bias stems from the skewed distribution of publicly available medical datasets, where certain imaging types dominate the training corpus while others remain severely undersampled. Consequently, models fail to generalize effectively across the full spectrum of clinical imaging modalities, limiting their applicability in comprehensive healthcare scenarios where practitioners routinely encounter diverse data types.

To address these challenges, we propose Fleming-VL, a unified medical MLLM that comprehensively understands 2D images, 3D volumetric scans, and temporal videos within a single end-to-end framework. Our approach tackles the aforementioned limitations from both model architecture and data perspectives. At the model level, we establish a unified data format by transforming all medical visual modalities into synthesized RGB representations, enabling seamless processing of heterogeneous inputs through a shared vision encoder. This design eliminates the need for modality-specific encoders while preserving the essential visual information across different data types. Additionally, we employ a two-stage training paradigm combining supervised fine-tuning (SFT) and group relative policy optimization (GRPO). SFT rapidly injects medical knowledge and task-specific capabilities, while GRPO enhances sampling efficiency and improves cross-modal robustness by optimizing the model's generation quality through preference learning. From the data perspective, we implement a comprehensive multi-stage training strategy that systematically addresses data scarcity and imbalance issues. During pretraining, we scale up the training corpus by integrating natural image interleaved data and medical image-caption pairs. The interleaved data extends the model's context window capacity, mitigating the distribution gap between short pretraining sequences and long contexts encountered during fine-tuning, while medical caption data facilitates domain-specific knowledge injection. In the fine-tuning stage, beyond curating existing open-source datasets, we introduce three key data enhancements. First, we augment chain-of-thought (CoT) reasoning data to address the shortcut learning problem prevalent in multiple-choice medical questions, encouraging the model to develop explicit reasoning processes rather than exploiting superficial patterns. Second, we develop an automated data production pipeline for underrepresented 2D modalities such as ultrasound and dermoscopy, generating high-quality instruction-tuning samples that significantly improve performance on these rare imaging types. Third, we construct and open-source a holistic video understanding dataset, filling the critical gap in medical video data availability. This dataset encompasses surgical procedures and general medical scenarios, with annotations including detailed descriptions and visual question answering pairs.

The effectiveness of Fleming-VL is comprehensively validated through extensive experiments across diverse medical imaging scenarios, as shown in Figure \ref{fig:scatter-1}. Our model achieves state-of-the-art performance on 9 benchmarks spanning 2D images, 3D volumes, and video modalities, demonstrating strong capabilities across a wide spectrum of clinical tasks including medical report generation, 3D image question answering, disease prediction, fine-grained video detail analysis, and holistic video content understanding. Notably, Fleming-VL exhibits superior cross-modal balance, avoiding the performance bias toward dominant modalities that plagues existing approaches. For instance, on the OmniMedVQA benchmark, our 8B model achieves 87\% accuracy, surpassing the significantly larger LingShu 32B model, highlighting the efficiency gains from our unified architecture and balanced training strategy. These results demonstrate that Fleming-VL not only advances the state-of-the-art in medical visual understanding but also provides a more practical and deployable solution for real-world clinical applications where diverse imaging modalities must be processed seamlessly within a single system.

\section{Related Work}

\subsection{Medical Multimodal Large Language Models}
The rapid emergence of large language models (LLMs) has profoundly accelerated advancements in medical artificial intelligence (AI). A growing number of domain-adapted medical LLMs have been proposed to enhance biomedical comprehension, clinical reasoning, and knowledge-grounded dialogue~\cite{wang2023huatuo,chen2024huatuogpt,flemingr1}. Building upon this foundation, the field has recently witnessed the rise of Medical Multimodal Large Language Models (MLLMs), which aim to integrate both textual and visual modalities for comprehensive medical understanding.

Early pioneering works such as LLaVA-Med, ChatDoctor, and PubMedGPT~\cite{li2023llava,li2023chatdoctor,venigalla2022pubmed} adapted general-purpose LLMs through fine-tuning on biomedical corpora and a limited number of image–text pairs. Despite these efforts, early MLLMs faced substantial challenges in modality alignment—they tended to over-rely on textual inputs while underutilizing visual information. This imbalance often led to hallucinations or inconsistencies when interpreting medical images, highlighting the need for more robust multimodal fusion strategies and clinically grounded training data.

To address these limitations, recent efforts have focused on preference optimization and modality alignment. MMedPO~\cite{zhu2024mmedpo} introduces two types of dispreference samples with clinically relevant weights to mitigate misalignment, while EXGRA-MED~\cite{nguyen2024logra} employs long-context graph alignment to strengthen semantic consistency across modalities. Complementary approaches such as MedGraphRAG, TrialGPT, and Med-PaLM 2~\cite{wu2024medical,jin2024matching,singhal2025toward} enhance factual grounding through retrieval-augmented reasoning and task-specific fine-tuning.

The latest generation of medical MLLMs emphasizes scale, comprehensive multimodal integration, and clinical reliability. Med-Gemini~\cite{medgemini2024} integrates long-context reasoning across textual, visual, and structured medical data for complex clinical decision-making. Lingshu~\cite{xu2025lingshu} establishes a unified foundation model supporting multiple medical imaging modalities, while MedVLM-R1~\cite{pan2025medvlm} advances interpretable visual reasoning through reinforcement learning on large-scale radiology–text corpora.

\subsection{Volumetric and Temporal Medical Understanding}
Recent research has extended medical MLLMs beyond standard 2D imaging to three-dimensional volumetric data and video modalities, enabling spatial reasoning and temporal understanding crucial for clinical and surgical applications. M3D~\cite{bai2024m3d} pioneers large-scale 3D multimodal learning, demonstrating integrated localization, segmentation, and report generation directly from volumetric scans. OmniV-Med~\cite{jiang2025omniv} unifies 2D, 3D, and video processing within a single framework, supporting multi-resolution data and both static and dynamic diagnostic tasks. Building on these foundations, Med3DVLM~\cite{xin2025med3dvlm} advances 3D vision–language modeling with strong cross-modal reasoning capabilities across retrieval, reporting, and question answering.

In the surgical domain, several works focus on understanding complex procedural workflows. SurgLLM~\cite{chen2025surgllm} emphasizes spatial–temporal reasoning for surgical procedures, enhancing multimodal understanding across captioning and question-answering tasks. SurgVidLM~\cite{wang2025surgvidlm} extends this direction toward multi-grained comprehension, bridging global procedural context with fine-grained action-level analysis. LLaVA-Surg~\cite{li2024llava} explores conversational and interactive reasoning over surgical videos, demonstrating strong capabilities in open-ended understanding and dialogue.

Collectively, these advances represent significant progress toward unified multimodal MLLMs that integrate volumetric and temporal comprehension for comprehensive clinical and surgical intelligence.

\section{Method}

\begin{figure}[t]
    \centering
        \includegraphics[width=\textwidth]{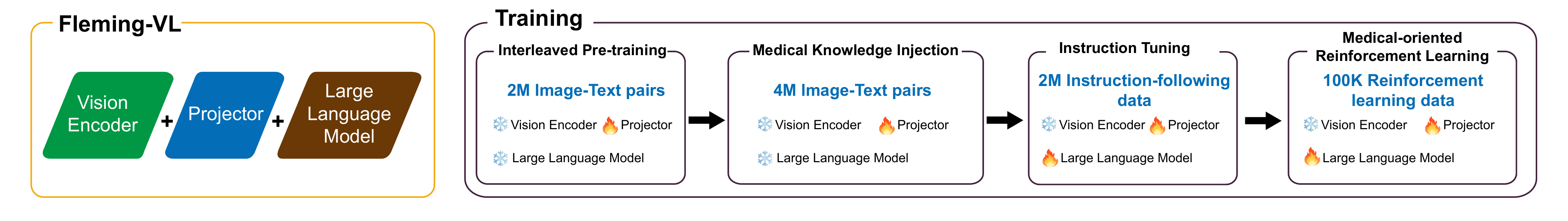}
        \caption{Fleming-VL data curation pipeline.}
        \label{fig:training}
        \vspace{-20pt}
\end{figure}

\subsection{Model Architecture}

Fleming-VL adopts a modular ``Vision-Projection-Language'' architecture that integrates visual and linguistic processing for medical image understanding. The model comprises three core components: a vision encoder for extracting visual features, a projection module for cross-modal alignment, and a language model for reasoning and generation. All components are initialized from InternVL3 \cite{zhu2025internvl3} pre-trained weights.

\textbf{Vision Encoder.} We employ InternViT as our vision encoder to extract visual representations from medical images. InternViT is a vision transformer designed for large-scale visual understanding tasks, available in two configurations: InternViT-300M and InternViT-6B. To efficiently process high-resolution medical images, the encoder incorporates a pixel unshuffle operation that reduces the visual token sequence length to one-quarter of its original size. Specifically, each $448 \times 448$ image tile is encoded into 256 visual tokens, enabling the model to handle detailed medical imagery while maintaining computational efficiency.

\textbf{Projection Module.} A two-layer multilayer perceptron (MLP) serves as the projection bridge connecting the vision and language modalities. This lightweight adapter transforms visual embeddings from the vision encoder into a representation space compatible with the language model, facilitating seamless cross-modal information flow.

\textbf{Language Model.} The linguistic backbone is built upon large language models (LLMs), specifically utilizing architectures from the Qwen2.5. The language model processes both projected visual tokens and text inputs through a unified autoregressive framework, generating medically-informed responses by attending to multimodal context.

\textbf{Position Encoding Strategy.} To accommodate extended medical imaging contexts—such as multi-view radiographs, temporal scan sequences, or whole-slide pathology images—Fleming-VL adopts Variable Visual Position Encoding (V2PE). Unlike traditional position encoding that assigns uniform increments to all tokens, V2PE employs modality-specific position increments: standard increments of 1 for text tokens and smaller fractional increments ($\delta < 1$) for visual tokens. This asymmetric encoding strategy compresses the positional space occupied by visual content, enabling the model to process longer multimodal sequences without exceeding the maximum context window, which is particularly valuable for analyzing multiple medical images simultaneously.

\subsection{Training}
 The training paradigm of Fleming-VL can be seen in Figure \ref{fig:training}, in which we discuss the details below.

\textbf{Interleaved Pre-training.} The first stage focuses on developing general visual-language reasoning abilities using interleaved image-text corpora. Unlike conventional vision-language pre-training that relies primarily on image-caption pairs, interleaved datasets contain documents with naturally interspersed images and text, more closely resembling real-world multimodal content such as medical reports, clinical notes, and educational materials.

We utilize 2M interleaved visual-language corpora \cite{zhu2023multimodal}  to enhance the model's ability to process and reason over complex multimodal contexts. This type of data is particularly valuable for medical applications, as clinical documentation often involves multiple images (e.g., different scan views, temporal series) accompanied by detailed textual descriptions. Training on interleaved data enables Fleming-VL to understand relationships across multiple images and their associated text, developing stronger contextual reasoning capabilities that are essential for comprehensive medical image interpretation.

\textbf{Medical Knowledge Injection.} The second stage establishes effective alignment between medical imaging modalities and their corresponding textual descriptions. This stage is crucial for adapting the general visual-language capabilities acquired in the first stage to the specialized medical domain. We utilize 4M medical image-caption datasets \cite{siragusa2024medpix,luo2024fairclip,seyfioglu2024quilt,subramanian2020medicat} to fine-tune the vision encoder and projection module, while keeping the language model frozen.

The medical caption data employed in this stage includes datasets with concise annotations that capture key diagnostic findings and anatomical observations. The relatively low semantic complexity of these captions facilitates rapid learning of general characteristics across diverse medical imaging modalities, including radiography, computed tomography, magnetic resonance imaging, and pathology slides. This not only aids in adjusting the vision encoder's outputs for better alignment with the language model but also contributes to faster convergence and improved training stability.

Through this two-stage pre-training paradigm, Fleming-VL develops a robust foundation for medical image understanding, combining general visual reasoning capabilities with specialized medical knowledge while maintaining strong linguistic competencies.

\textbf{Instruction Tuning.} Following the pre-training stages, we conduct instruction tuning to refine Fleming-VL's ability to follow diverse clinical task directives and align its outputs with medical practice requirements. In accordance with established practices in multimodal large language models, this stage is designed to enhance the model's instruction-following capabilities across a broad spectrum of medical scenarios. To accommodate the diverse requirements of clinical applications, we unlock all model parameters and perform large-scale, end-to-end optimization of the entire architecture.

The instruction tuning corpus extends beyond conventional formats such as image descriptions, question-answer pairs, and multiple-choice questions. We curate a comprehensive dataset comprising approximately 2 million instruction-response pairs, integrating an extensive collection of scenario-oriented queries derived from real clinical workflows and multiple authoritative medical datasets, including PathVQA \cite{he2020pathvqa}, PMC-VQA \cite{zhang2023pmc}, SLAKE \cite{liu2021slake}, Quilt-LLaVA \cite{seyfioglu2024quilt}, VQA-Med-2019 \cite{ben2019vqa}, MIMIC-CXR-VQA \cite{bae2024mimic}, PubMedVision \cite{chen2024huatuogpt}, LLaVA-Med \cite{li2023llava}, VQA-RAD \cite{lau2018dataset}, MIMIC-CXR \cite{johnson2019mimic}, and IU-Xray \cite{demner2015preparing}. These queries encompass diverse clinical tasks including differential diagnosis, radiological examination interpretation, medical knowledge retrieval, clinical report generation, and anatomical structure localization. This comprehensive coverage substantially enhances the model's competency across various medical domains.

However, existing open-source medical multimodal datasets exhibit two significant limitations. First, there exists a pronounced modality bias: while X-ray and CT/MRI images are abundantly represented, other critical imaging modalities such as ultrasound, dermoscopy, and ophthalmology remain severely underrepresented. Second, the available data is predominantly image-centric, with a notable scarcity of medical video data despite the increasing importance of dynamic imaging in clinical practice (e.g., echocardiography, surgical videos, endoscopy procedures).

To address these data imbalances, we synthesize a substantial volume of instruction tuning samples targeting underrepresented modalities and video-based medical content. This data synthesis strategy, which will be detailed in the next section, enables Fleming-VL to achieve more balanced multimodal coverage and develop stronger capabilities in processing temporal medical information. The resulting diversified instruction tuning regimen, encompassing 2 million high-quality instruction-response pairs, equips Fleming-VL to navigate complex clinical scenarios across a wide spectrum of imaging modalities with enhanced precision and reliability.

\begin{figure}[t]
    \centering
        \includegraphics[width=\textwidth]{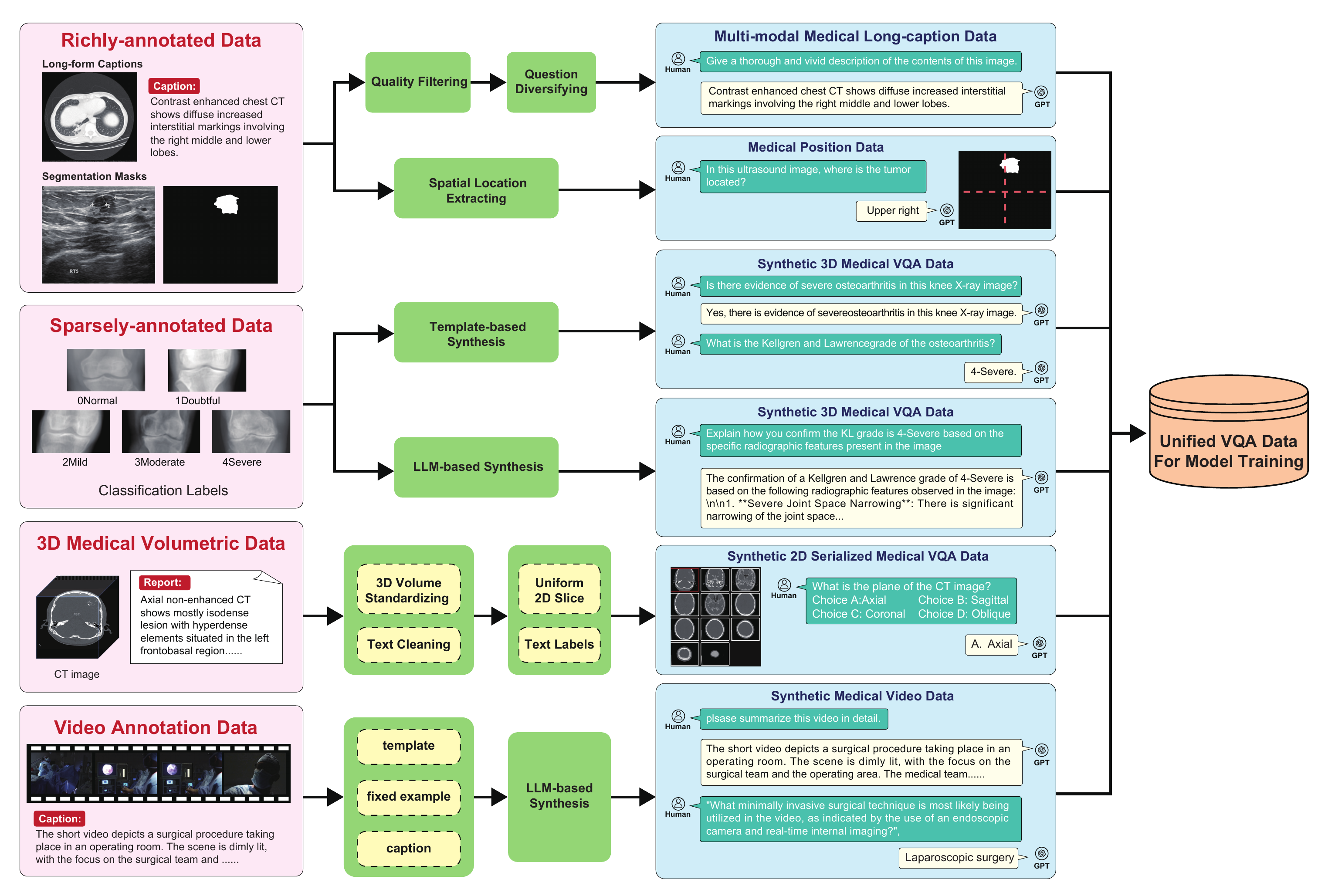}
        \caption{Fleming-VL data curation pipeline.}
        \label{fig:data_curation}
        \vspace{-20pt}
\end{figure}

\textbf{Medical-oriented Reinforcement Learning.} Recent breakthroughs in reasoning models, exemplified by OpenAI's o-series and DeepSeek-R1~\cite{guo2025deepseek}, have demonstrated the effectiveness of reinforcement learning with verifiable rewards (RLVR) for enhancing complex reasoning capabilities. Group Relative Policy Optimization (GRPO)~\cite{shao2024deepseekmath} has emerged as a particularly effective approach and has been increasingly adopted for multimodal large language models. Conventional supervised fine-tuning encounters critical limitations in medical applications: excessive reliance on answer supervision can lead to overfitting and shortcut learning, where models memorize surface patterns rather than developing genuine clinical reasoning capabilities. Reinforcement learning addresses these challenges by incentivizing models to autonomously discover robust reasoning pathways through reward signals.

Following this paradigm, we adopt GRPO to enhance Fleming-VL's clinical reasoning and generalization capabilities using a carefully curated dataset of 100K samples randomly drawn from the instruction tuning corpus. To construct verifiable training signals, we reformulate multiple-choice questions into open-ended format and downsample binary yes/no questions to approximately 5\% to mitigate bias. For reward design, we employ format-based rewards and accuracy-based rewards tailored to different question formats: direct option matching (A, B, C, D) for multiple-choice questions, and an LLM-as-judge approach for open-ended questions that assesses response correctness against reference answers. This reinforcement learning stage refines Fleming-VL's systematic clinical reasoning capabilities, ultimately enhancing its reliability in real-world medical applications.

\subsection{Data Curation}

As discussed in Section 3.2, existing open-source medical multimodal datasets exhibit two critical limitations. First, modality bias: X-ray and CT/MRI images dominate available data, while clinically important modalities such as ultrasound, dermoscopy, and ophthalmology remain severely underrepresented. Second, temporal data scarcity: the predominance of static images with minimal medical video data limits models' ability to interpret dynamic clinical content such as echocardiography and endoscopy.

To address these imbalances, we implement a comprehensive data curation strategy targeting underrepresented modalities and video content, as shown in Figure \ref{fig:data_curation}. Through systematic collection from multiple public sources and automated synthesis pipelines, we construct a large-scale, balanced medical multimodal dataset that enables Fleming-VL to achieve equitable coverage across imaging modalities and develop stronger temporal reasoning capabilities. Given the heterogeneous annotation formats across public datasets, we implement a systematic synthesis pipeline to transform raw medical data into high-quality visual question-answering formats. Our strategy comprises four core approaches:

\paragraph{VQA Conversion for Richly-annotated Data.}
For datasets with detailed descriptions or segmentation masks, we perform structured VQA conversion. Image-caption pairs are reformulated as question-answer pairs with diversified questions sampled from synonym pools to enhance linguistic variability. For segmented datasets, we algorithmically extract pixel coordinates to generate spatial localization questions (e.g., "Where is the lesion?" → "Left upper lobe"). All samples undergo quality filtering and are prepared in bilingual Chinese-English versions.

\paragraph{Synthesis for Sparse Data.}
For sparsely annotated datasets with only classification labels, we employ two methods. First, template-based generation: creating multiple-choice questions with ground-truth labels as answers and random labels as distractors, or direct diagnostic questions for open-ended formats. Second, LLM-assisted synthesis: leveraging GPT-4o with raw images, diagnostic labels, lesion characteristics, patient metadata, and domain knowledge to generate detailed, reasoning-oriented question-answer pairs. We constrain generation to visual content and verified annotations, ensuring factual reliability while enriching simple labels into interpretable instructions. We organize the data source in Table \ref{tab:datasets}.

\paragraph{3D Volumetric Data Processing.}
For 3D medical imaging datasets, we implement a slice-based serialization approach to adapt volumetric data for multi-image processing. Each 3D volume is converted into a sequence of 2D RGB images through evenly-spaced sampling along the depth dimension. These serialized slices are treated as an ordered image sequence, with the original 3D annotations transformed into sequential VQA samples. This approach enables the model to interpret 3D anatomical structures through multi-image reasoning while maintaining compatibility with the standard 2D image processing pipeline.

\paragraph{Medical Video Data Synthesis.}
For medical video datasets, we develop a synthesis pipeline to enhance the model's video understanding capabilities, particularly in capturing temporal dynamics and fine-grained visual details. We leverage MedVideoCap \cite{wang2025medgen}, a medical video-caption dataset, as the foundation. Each sample consists of a raw video paired with its descriptive caption, which we utilize as ground-truth labels for video summarization tasks (e.g., "Describe this medical procedure" → {original caption}). To further assess the model's comprehension of specific video content and temporal details, we employ GPT-4o to generate multiple-choice questions based on these captions. The generation prompt comprises two components: a structured response template specifying the desired output format (one medically appropriate question, one correct answer, and 1-3 distractors), and a few-shot example demonstrating the expected quality. This approach produces fine-grained questions that require careful attention to procedural steps, anatomical landmarks, and temporal sequences within the video, thereby evaluating the model's ability to extract and reason about detailed visual information rather than providing only high-level descriptions.

Through this curation process, we construct a high-quality medical multimodal VQA dataset comprising 2M samples across diverse modalities and clinical scenarios, forming the foundation for Fleming-VL's robust instruction tuning and enhanced generalization.

\begin{table}[t]
\centering
\small
\caption{List of medical imaging datasets of different modalities.}
\label{tab:datasets}  

\begin{tabularx}{\linewidth}{lXX}  
\toprule
\textbf{Modality} & \textbf{Collected Datasets} & \textbf{Description}\\
\midrule
X-ray &  Digital Knee X-ray \cite{gornale2020comprehensive} & Knee X-ray images \& labels\\
\addlinespace[0.3em] 
\hline
\addlinespace[0.3em] 
CT & M3D \cite{bai2024m3d} & 3D medical volumes \& QA pairs\\
\addlinespace[0.3em] 
\hline
\addlinespace[0.3em] 
MRI & Brain-Tumor-MRI \cite{msoud_nickparvar_2021} & 2D MRI slices \& free-form reports\\
\addlinespace[0.3em] 
\hline
\addlinespace[0.3em] 
\multirow{3}{*}{Ultrasound} & AbdomenUS \cite{vitale2020improving} & Abdominal ultrasound images \& lesion segmentation masks \\
& Breast Ultrasound Images \cite{ALDHABYANI2020104863} & Breast ultrasound images \& labels \\
& Annotated Ultrasound Liver images \cite{xu2022annotated} & Liver ultrasound images \& lesion segmenta-
tion masks \\

\addlinespace[0.3em] 
\hline
\addlinespace[0.3em] 
Dermoscopy & PAD-UFES-20 \cite{pacheco2020pad} & Skin lesion images \& labels\\
\addlinespace[0.3em] 
\hline
\addlinespace[0.3em] 
Ophthalmology & EyePACS \cite{cuadros2009eyepacs} & Fundus images \& labels \\
\addlinespace[0.3em] 
\hline
\addlinespace[0.3em] 
\multirow{3}{*}{Multimodal} 
 & ROCOv2 \cite{ruckert2024rocov2} & Radiology image \& caption pairs from publications \\
 & Harvard-FairVLMed \cite{luo2024fairclip} & Fundus images \& clinical notes\\
 & MedICaT \cite{subramanian2020medicat} & Medical publication figures \& caption pairs\\
 & MedPix-2.0 \cite{siragusa2024medpix} & Medical knowledge base with images \& cases\\
\bottomrule
\end{tabularx}
\end{table}


\section{Implementation Details}


\textbf{Stage 1 \& 2: Pre-training and Knowledge Injection.}  We use a learning rate of 1e-5 with cosine annealing scheduler, 3\% linear warmup, and weight decay of 0.05. The effective batch size is 128 across 8 GPUs with gradient accumulation steps of 2. Dynamic image resizing is enabled to handle varying input resolutions up to 448$\times$448 pixels. Both stages are trained for 1 epoch using the AdamW optimizer ($\beta_1$=0.9, $\beta_2$=0.999, $\epsilon$=1e-8).

\textbf{Stage 3: Instruction Tuning.} 
We increase the learning rate to 2e-5 while maintaining other hyperparameters identical to the previous stages. 

\textbf{Stage 4: Medical-oriented Reinforcement Learning.} 
We adopt Group Relative Policy Optimization (GRPO)~\cite{shao2024deepseekmath} for reinforcement learning with 8 generations per prompt (N=8) at temperature 0.9. The KL divergence penalty coefficient $\beta$ is set to 0.04 to balance exploration and exploitation. Following~\cite{shao2024deepseekmath}, we use a significantly lower learning rate of 1e-6 with no weight decay, 10\% warmup ratio, and disable dynamic image sizing for training stability. The data is divided into 4 groups for mini-batch updates within each epoch. We train for 1 epoch with gradient clipping at max norm 1.0.

\begin{table*}[t]
\centering
\caption{Training Hyperparameters for Four-Stage Medical VLM Fine-tuning}
\label{tab:exp_settings}

\begin{tabular}{lcccc}
\toprule
\textbf{Hyperparameter} & \textbf{Stage 1} & \textbf{Stage 2} & \textbf{Stage 3} & \textbf{Stage 4} \\
 & \textbf{Interleaved} & \textbf{Medical Knowledge} & \textbf{Instruction} & \textbf{Medical-oriented} \\
 & \textbf{Pre-training} & \textbf{Injection} & \textbf{Tuning} & \textbf{RL} \\
\midrule
BF16 & True & True & True & True \\
Learning Rate & 1e-5 & 1e-5 & 2e-5 & 1e-6 \\
Weight Decay & 0.05 & 0.05 & 0.05 & 0.0 \\
LR Scheduler & cosine & cosine & cosine & cosine \\
Warmup Ratio & 0.03 & 0.03 & 0.03 & 0.1 \\
Dynamic Image Size & True & True & True & False \\
\midrule
\multicolumn{5}{l}{\textit{GRPO Specific Parameters}} \\
\midrule
Num Generations ($N$) & - & - & - & 8 \\
Temperature & - & - & - & 0.9 \\
Beta ($\beta$) & - & - & - & 0.04 \\
Num of Group & - & - & - & 4 \\
\bottomrule
\end{tabular}
\end{table*}

\section{Experiments}
We evaluated the model’s performance using multimodal QA and report-generation-style QA. Simultaneously, we constructed benchmark tasks on video and 3-D datasets to assess the model’s video-understanding and 3-D image-understanding capabilities.

\subsection{Evaluation Benchmarks}

To comprehensively evaluate Fleming-VL's medical capabilities, we select a diverse suite of established medical vision-language benchmarks covering multiple imaging modalities, question formats, and clinical domains:

\textbf{OmniMedVQA}~\cite{hu2024omnimedvqa} is a large-scale collection spanning 12 medical imaging modalities including CT, MRI, X-ray, ultrasound, and pathology. We evaluate on eight representative modalities to assess cross-modal generalization.

\textbf{PMC-VQA}~\cite{zhang2023pmc} contains 227K question-answer pairs over 149K images extracted from PubMed Central articles, covering diverse pathologies and imaging modalities with both open-ended and multiple-choice questions.

\textbf{VQA-RAD}~\cite{lau2018dataset} is an expert-curated dataset comprising 315 radiology images with approximately 10 question-answer pairs per image, focusing on diagnostic reasoning in radiological contexts.

\textbf{PathVQA}~\cite{he2020pathvqa} comprises 32,799 questions generated from 4,998 pathology images extracted from medical textbooks, evaluating the model's understanding of histopathological features and microscopic structures.

\textbf{SLAKE}~\cite{liu2021slake} is a bilingual (English-Chinese) dataset with semantically rich annotations from experienced clinicians, featuring a structured medical knowledge base that enables evaluation of both visual understanding and medical knowledge reasoning.

\textbf{MIMIC-CXR}~\cite{johnson2019mimic} is a large-scale chest X-ray dataset containing 377,110 images from 227,835 radiographic studies, each accompanied by free-text radiology reports. We evaluate medical report generation by assessing the model's ability to produce structured clinical descriptions covering findings, impressions, and diagnostic interpretations.

\textbf{IU-Xray}~\cite{demner2015preparing} comprises 7,470 chest X-ray images paired with diagnostic reports from Indiana University Hospital. This benchmark focuses on generating coherent radiology reports that capture key clinical observations and pathological findings from frontal and lateral chest radiographs.

\textbf{M3D-VQA}~\cite{bai2024m3d} is a 3D medical imaging benchmark with 2,000 volumetric images and 13,791 multiple-choice questions covering five question types. It evaluates the model's ability to interpret complex 3D anatomical structures and spatial relationships in volumetric medical data.

\textbf{MedVideoBench} is our proposed medical video understanding benchmark comprising raw-format medical videos with two evaluation tasks: holistic video understanding and multiple-choice VQA in visual content, assessing the model's ability to capture temporal dynamics and procedural details in medical videos. For holistic video understanding, we employ ROUGE-L and CIDEr metrics to evaluate the quality and clinical fidelity of generated video summaries, while multiple-choice questions are evaluated using accuracy.

These benchmarks collectively assess Fleming-VL's performance across key dimensions including diagnostic accuracy, cross-modal generalization, multilingual understanding, and medical knowledge integration.

\begin{table}[t]
\centering
\caption{General medical vqa benchmarks performance. All scores are scaled by a factor of 100 to enhance clarity and comprehension.}
\label{tab:results}
\setlength{\tabcolsep}{3mm}   
\renewcommand{\arraystretch}{1.1} 
\adjustbox{max width=\linewidth,center}{%
\begin{tabular}{lcccccc}
\toprule
\textbf{Model} & \textbf{OMVQA} & \textbf{PMC-VQA} & \textbf{PathVQA} & \textbf{Slake} & \textbf{VQA-RAD} & \textbf{Avg.} \\
\midrule
\multicolumn{7}{c}{\texttt{Proprietary Models}} \\
\midrule
Claude Sonnet 4     & 65.5 & 54.4 & 54.2 & 70.6 & \underline{67.6} & 62.5 \\
GPT-5             & \textbf{77.9} & \textbf{58.3} & \textbf{56.2} & \underline{73.4} & 66.2 & \textbf{66.4} \\
Gemini-2.5-Flash    & \underline{71.0} & \underline{55.4} & \underline{55.4} & \textbf{75.8} & \textbf{68.5} & \underline{65.2} \\
\midrule
\multicolumn{7}{c}{\texttt{Open-source Models (<10B)}} \\
\midrule
BiomedGPT         & 27.9 & 27.6 & 11.3 & 13.6 & 16.6 & 19.4 \\
LLaVA-Med-7B      & 44.3 & 30.5 & 38.8 & 48.0 & 53.7 & 43.1 \\
MedVLM-R1-2B      & 77.7 & 47.6 & 32.5 & 56.0 & 48.6 & 52.5 \\
BioMediX2-8B      & 63.3 & 43.5 & 37.0 & 57.7 & 49.2 & 50.1 \\
Med-R1-2B         & --   & 47.4 & 15.3 & 54.5 & 39.0 & --   \\
Qwen2.5VL-7B      & 63.6 & 51.9 & 44.1 & 67.2 & 64.5 & 58.2 \\
InternVL2.5-8B    & 81.3 & 51.3 & 42.1 & 69.0 & 59.4 & 60.6 \\
HuatuoGPT-V-7B    & 74.2 & 53.3 & 48.0 & 67.8 & 67.0 & 62.1 \\
MedGemma-4B-IT    & 69.8 & 49.9 & 48.8 & 76.4 & \textbf{72.5} & 63.5 \\
InternVL3-8B      & 79.1 & 53.8 & 48.6 & 72.8 & 65.4 & 64.0 \\
Qwen3VL-8B-thinking     & 77.1 & 54.3 & 47.8 & 71.8 & 64.3 & 63.1 \\
Lingshu-7B        & \underline{82.9} & \underline{56.3} & \underline{61.9} & \underline{83.1} & \underline{67.9} & \underline{70.4} \\
\rowcolor{ModelGreen}\textbf{Fleming-VL-8B} & \textbf{86.7} & \textbf{64.3} & \textbf{62.9} & \textbf{86.5} & 66.1 & \textbf{73.3} \\
\midrule
\multicolumn{7}{c}{\texttt{Open-source Models (>10B)}} \\
\midrule
Qwen2.5V-32B      & 68.2 & 54.5 & 41.9 & 71.2 & 71.8 & 61.5 \\
InternVL2.5-38B   & 79.9 & 57.2 & 46.9 & 70.3 & 61.4 & 63.1 \\
InternVL3-14B     & 78.9 & 54.1 & 48.0 & 72.8 & 66.3 & 64.0 \\
InternVL3-38B     & 79.8 & 56.6 & 51.0 & 72.7 & 65.4 & 65.1 \\
Lingshu-32B        & \underline{83.4} & \underline{57.9} & \underline{65.9} & \underline{89.2} & \underline{76.5} & \underline{74.6} \\
\rowcolor{ModelGreen}\textbf{Fleming-VL-38B} & \textbf{87.9} & \textbf{76.5} & \textbf{68.0} & \textbf{89.8} & \textbf{76.6} & \textbf{79.8} \\
\bottomrule
\end{tabular}}
\end{table}

\begin{table}[t]
\centering
\caption{Comprehensive evaluation of medical report generation on MIMIC-CXR and IU-Xray. All scores are scaled by a factor of 100 to enhance clarity and comprehension.}
\adjustbox{max width=1.0\textwidth}{
\setlength{\tabcolsep}{0.8mm}{
\begin{tabular}{lcccccccccc}
\toprule
\multirow{3}{*}{\textbf{Models}}& \multicolumn{5}{c}{\textbf{MIMIC-CXR}} & \multicolumn{5}{c}{\textbf{IU-Xray}} \\
\cmidrule(lr){2-6} \cmidrule(lr){7-11}
 & \textbf{ROUGE-L} & \textbf{CIDEr} & \textbf{RaTE}  & \textbf{Semb}  & \textbf{RadCliQ$^{-1}$} & \textbf{ROUGE-L} & \textbf{CIDEr} & \textbf{RaTE}  & \textbf{Semb}  & \textbf{RadCliQ$^{-1}$}\\
\midrule
\multicolumn{11}{c}{\texttt{Proprietary Models}}\\
\midrule
GPT-4.1 & 9.0 & \textbf{82.8} & \textbf{51.3} & \underline{23.9} & \underline{57.1} & \underline{30.2}  & \underline{124.6} & 51.3 & \underline{47.5} & \underline{80.3} \\
Claude Sonnet 4 & \underline{20.0} & 56.6 & 45.6 & 19.7 & 53.4 & 25.4 & 88.3 & \underline{55.4} & 41.0 & 72.1 \\
Gemini-2.5-Flash & \textbf{25.4} & \underline{80.7} & \underline{50.3} & \textbf{29.7} & \textbf{59.4} & \textbf{33.5} & \textbf{129.3} & \textbf{55.6} & \textbf{50.9} & \textbf{91.6} \\
\midrule
\multicolumn{11}{c}{\texttt{Open-source Models (<10B)}}\\
\midrule
Med-R1-2B & 19.3 & 35.4 & 40.6 & 14.8 & 42.4 & 16.1 & 38.3 & 41.4 & 12.5 & 43.6 \\
MedVLM-R1-2B & 20.3 & 40.1& 41.6& 14.2 & 48.3 & 22.7 & 61.1 & 46.1 & 22.7 & 54.3\\
MedGemma-4B-IT & 25.6 & 81.0 & \underline{52.4} & 29.2 & 62.9 & 30.8 & 103.6 & 57.0 & 46.8 & 86.7\\
LLaVA-Med-7B & 15.0 & 43.4 & 12.8 & 18.3 & 52.9 & 18.8 & 68.2 & 40.9 & 16.0 & 58.1\\
HuatuoGPT-V-7B & 23.4 & 69.5 & 48.9 & 20.0 & 48.2 & 29.6 & 104.3 & 52.9 & 40.7 & 63.6 \\
BioMediX2-8B & 20.0  & 52.8 & 44.4 & 17.7 & 53.0 & 19.6  & 58.8 & 40.1 & 11.6 & 53.8\\
Qwen2.5VL-7B & 24.1 & 63.7 & 47.0  & 18.4 & 55.1 & 26.5 & 78.1 & 48.4  & 36.3 & 66.1 \\
InternVL2.5-8B   & 23.2 & 61.8 & 47.0 &21.0 & 56.2 & 24.8 & 75.4 & 51.1 & 36.7 & 67.0 \\
InternVL3-8B & 22.9 & 66.2 & 48.2 &21.5 & 55.1 & 22.9  & 76.2 & 51.2 & 31.3 & 59.9 \\
Lingshu-7B & \underline{30.8} & \underline{109.4} & 52.1  & \underline{30.0} & \underline{69.2} & \underline{41.2}  & \underline{180.7} & \underline{57.6} & \underline{48.4} & \underline{108.1}  \\
\rowcolor{ModelGreen}\textbf{Fleming-VL-8B} & \textbf{35.7} & \textbf{132.5} & \textbf{56.7} & \textbf{33.6} & \textbf{70.9} & \textbf{44.9} & \textbf{198.6} & \textbf{66.7} & \textbf{51.3} & \textbf{112.1} \\
\midrule
\multicolumn{11}{c}{\texttt{Open-source Models (>10B)}}\\
\midrule
HealthGPT-14B &  21.4 & 64.7 & 48.4 & 16.5 & 52.7 & 22.9 & 81.9 & 50.8 & 16.6 & 56.9 \\
HuatuoGPT-V-34B & 23.5 & 68.5 & 48.5 & 23.0 & 47.1 & 28.2 & 108.3 & 54.4 & 42.2 & 59.3 \\
MedDr-40B  & 15.7 & 62.3 & 45.2 & 12.2 & 47.0 & 19.4 & 62.9 & 40.3 & 7.3 & 48.9\\
InternVL3-14B & 22.0 & 63.7 & 48.6 & 17.4 & 46.5 & 24.8 & 93.7 & 55.0 & 38.7 & 55.0 \\
Qwen2.5VL-32B  & 15.7 & 50.2 & 47.5 & 17.1 & 45.2 & 18.9 & 73.3 & 51.3 & 38.1 & 54.0 \\
InternVL2.5-38B & 22.7 & 61.4 & 47.5 & 18.2 & 54.9 & 28.9 & 96.5  & 53.5 & 38.5 & 69.7 \\
InternVL3-38B & 22.8 & 64.6 & 47.9 & 18.1 & 47.2 & 25.5 & 90.7 & 53.5 & 33.1 & 55.2 \\
Lingshu-32B & \underline{28.8} & \underline{96.4} & \underline{50.8} & \underline{30.1} & \underline{67.1} & \underline{42.8} & \underline{189.2} & \underline{63.5} & \underline{54.6} & \underline{130.4} \\
\rowcolor{ModelGreen}\textbf{Fleming-VL-38B} & \textbf{35.9} & \textbf{133.6} & \textbf{56.9} & \textbf{33.7} & \textbf{71.2} & \textbf{43.2} & \textbf{190.0} & \textbf{63.9} & \textbf{55.8} & \textbf{132.2} \\
\bottomrule
\end{tabular}}}
\label{tab:report_result}
\end{table}

\begin{table}[t]
\centering
\small
\caption{Main results on video understanding and 3D benchmarks. Our model demonstrates strong performance on both holistic video understanding and medical video question answering tasks.}
\label{tab:video_results}
\begin{tabular}{lcccc}
\toprule
\multirow{2}{*}{\textbf{Model}} &
\multicolumn{2}{c}{\textbf{Holistic Video Understanding}} &
\textbf{Medical} &
\textbf{3D-Medical} \\
\cmidrule(lr){2-3}
& \textbf{ROUGE-L} & \textbf{CIDEr} & \textbf{Video VQA} & \textbf{VQA} \\
\midrule
\multicolumn{5}{c}{\texttt{Open-source Models (<10B)}} \\
\midrule
InternVL3-8B & 34.4 & 181.8 & 85.4 & 56.7 \\
Lingshu-7B & 33.8 & 178.7 & 63.2 & 56.0 \\
Qwen2.5-VL-7B & 35.6 & 188.8 & 87.4 & 51.2 \\
\rowcolor{ModelGreen}\textbf{Fleming-VL-8B} & \textbf{44.7} & \textbf{285.0} & \textbf{90.2} & \textbf{61.9} \\
\midrule
\multicolumn{5}{c}{\texttt{Open-source Models (>10B)}} \\
\midrule
InternVL3-38B & 35.6 & 192.3 & 87.2 & 57.3 \\
Qwen2.5-VL-32B & 34.3 & 185.6 & 66.8 & 56.5 \\
Lingshu-32B & 37.4 & 197.2 & 56.4 & 53.8 \\
\rowcolor{ModelGreen}\textbf{Fleming-VL-38B} & \textbf{44.9} & \textbf{287.3} & \textbf{91.6} & \textbf{59.6} \\
\bottomrule
\end{tabular}
\end{table}



\subsection{Empirical results and Analysis}\label{sec:results}

\paragraph{General Medical VQA Performance.}
Fleming-VL establishes new state-of-the-art results among open-source models across multiple medical VQA benchmarks at both model scales (Table~\ref{tab:results}). At the sub-10B scale, Fleming-VL-8B achieves a macro average of 73.0, significantly outperforming the previous best open-source model Lingshu-7B (73.3) by +3 points and InternVL3-8B (64.0) by +9 points. Notably, Fleming-VL-8B demonstrates strong performance across diverse datasets: 86.7 on OmniMedVQA, 64.3 on PMC-VQA, 62.9 on PathVQA, 86.5 on SLAKE, and 66.1 on VQA-RAD.

At the larger scale, Fleming-VL-38B achieves an average of 79.8, surpassing Lingshu-32B (74.6) by +5.2 points. The improvements are consistent across all benchmarks: 87.9 on OmniMedVQA (+4.5 over Lingshu-32B), 76.5 on PMC-VQA (+18.6), 68.0 on PathVQA (+2.1), 89.8 on SLAKE (+0.6), and 76.6 on VQA-RAD (+0.1). The particularly substantial gain on PMC-VQA demonstrates Fleming-VL's enhanced ability to handle diverse medical literature contexts and transfer domain knowledge effectively. These results indicate that our comprehensive data curation strategy and training methodology successfully improve both average performance and cross-modality generalization.

\paragraph{Radiology Report Generation.}
As shown on Table \ref{tab:report_result}, on MIMIC-CXR and IU-Xray benchmarks, Fleming-VL demonstrates substantial improvements across both semantic overlap and clinical fidelity metrics. At the sub-10B scale, Fleming-VL-8B surpasses Lingshu-7B on MIMIC-CXR across all metrics: ROUGE-L +4.9 (35.7 vs. 30.8), CIDEr +23.1 (132.5 vs. 109.4), RaTE +4.6 (56.7 vs. 52.1), Semb +3.6 (33.6 vs. 30.0), and RadCliQ$^{-1}$ +1.7 (70.9 vs. 69.2). On IU-Xray, Fleming-VL-8B achieves better performance compared to baselines.

At the larger scale, Fleming-VL-38B shows even more pronounced improvements over Lingshu-32B on MIMIC-CXR: ROUGE-L +7.1 (35.9 vs. 28.8), CIDEr +37.2 (133.6 vs. 96.4), RaTE +6.1, Semb +3.6, and RadCliQ$^{-1}$ +4.1. On IU-Xray, Fleming-VL-38B maintains consistent positive gains across all metrics, including CIDEr +0.8. These results demonstrate that Fleming-VL effectively generates clinically accurate and semantically coherent radiology reports, with particularly strong performance on the large-scale MIMIC-CXR benchmark reflecting improved generalization to diverse radiological findings.

\paragraph{Video and 3D Medical Understanding.}
Fleming-VL demonstrates substantial improvements in temporal and volumetric medical reasoning, particularly excelling in holistic video understanding tasks. At the sub-10B scale, Fleming-VL-8B achieves 44.7 ROUGE-L and 285.0 CIDEr on video summarization—representing improvements of +9.1 and +96.2 over Qwen2.5-VL-7B, and significantly outperforming both InternVL3-8B (34.4/181.8) and Lingshu-7B (33.8/178.7). For structured tasks, Fleming-VL-8B attains 90.2 on Medical Video-VQA and 0.619 on 3D-Medical VQA, establishing new state-of-the-art results among sub-10B models.

At the larger scale, Fleming-VL-38B continues this trend with 44.9 ROUGE-L and 287.3 CIDEr on video summarization, surpassing all open-source models including Lingshu-32B by +7.5 and +90.1 respectively. Notably, Fleming-VL-38B achieves 91.6 on Medical Video-VQA—substantially higher than InternVL3-38B (87.2) and Lingshu-32B (56.4)—demonstrating exceptional understanding of temporal medical procedures. The consistent gains across both video summarization and structured VQA tasks indicate that our targeted data synthesis for underrepresented modalities successfully bridges the gap between holistic temporal understanding and fine-grained visual reasoning. These results validate that addressing data scarcity in medical videos and 3D volumetric imaging through systematic curation and synthesis enables models to develop robust capabilities in processing dynamic and spatial medical information.

\begin{table}[h]
\centering
\caption{Ablation study on key training components. All experiments are conducted with Fleming-VL-8B.}
\label{tab:ablation}
\resizebox{\linewidth}{!}{%
\begin{tabular}{lccccccc}
\toprule
\textbf{Model} & \textbf{OMVQA} & \textbf{PMC-VQA} & \textbf{PathVQA} & \textbf{Slake} & \textbf{VQA-RAD} & \textbf{Video VQA} & \textbf{3D VQA} \\
\midrule
Fleming-VL-8B & 89.1 & 64.3 & 55.4 & 75.8 & 68.5 & 90.2 & 61.9 \\
w/o interleaved reasoning & 88.2 & 63.6 & 55.6 & 74.3 & 67.9 & 88.5 & 59.2 \\
w/o medical caption pretraining & 87.5 & 63.0 & 55.0 & 73.7 & 66.8 & 87.4 & 59.0 \\
w/o hard samples synthesis & 82.3 & 62.9 & 51.9 & 74.2 & 64.8 & 87.0 & 58.2 \\
\bottomrule
\end{tabular}%
}
\end{table}

\subsection{Ablation Study}

To validate the effectiveness of our key design choices, we conduct ablation experiments by systematically removing each component from Fleming-VL-8B and evaluating performance across seven medical understanding benchmarks (Table~\ref{tab:ablation}).

\textbf{Interleaved reasoning pretraining.} Removing interleaved image-text pretraining data leads to consistent performance drops across all benchmarks, with particularly notable degradation on Video VQA (-1.7) and 3D VQA (-2.7). This demonstrates that training on naturally interspersed multimodal content enhances the model's ability to reason across multiple images and understand complex spatial-temporal relationships, which directly transfers to video and volumetric understanding tasks.

\textbf{Medical caption pretraining.} Ablating the medical image-caption alignment stage results in decreased performance across all tasks, with larger drops on video understanding (-2.8) and 3D reasoning (-2.9). This validates that establishing effective vision-language alignment through domain-specific medical captions is crucial for adapting general visual capabilities to specialized medical contexts, particularly for complex modalities.

\textbf{Hard samples synthesis.} The removal of synthesized samples targeting underrepresented modalities causes the most substantial performance degradation, particularly on OMVQA (-6.8) and 3D VQA (-3.7). This significant impact underscores the critical importance of our data curation strategy in addressing modality imbalance. Without targeted synthesis for rare modalities such as ultrasound, dermoscopy, and 3D volumetric data, the model's ability to generalize across the full spectrum of medical imaging is substantially compromised.





\begin{figure}[t]
    \centering
    \includegraphics[width=0.98\textwidth]{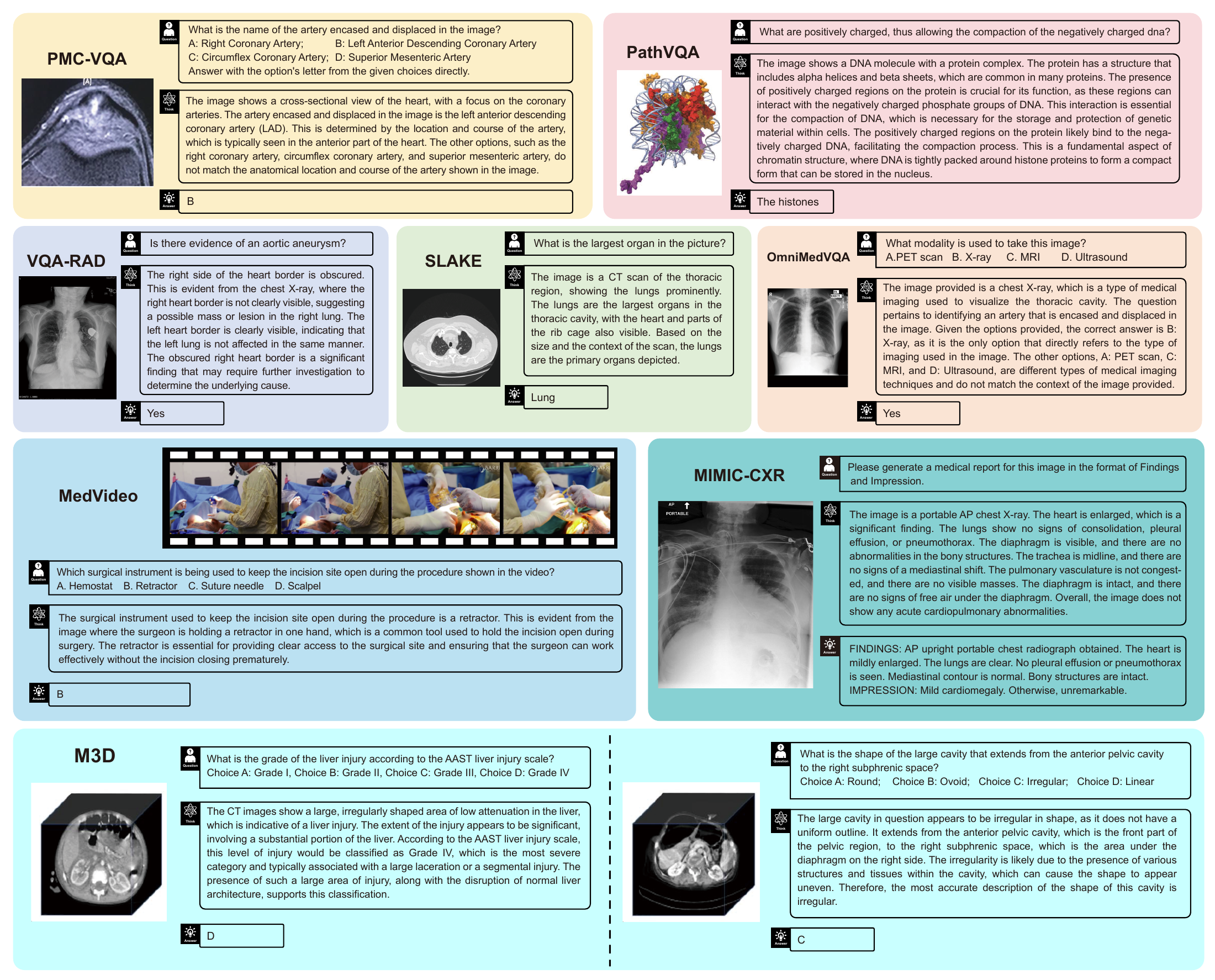}
    \caption{Visualization of Fleming-VL,in which models can reason across different modalities.}
    \label{fig:vis}
\end{figure}

\section{Visualization}
Figure \ref{fig:vis} demonstrates Fleming-VL's comprehensive capabilities across diverse medical imaging modalities and tasks. The model processes various input types including 2D radiological images (VQA-RAD, MIMIC-CXR), 3D volumetric scans (M3D), pathology slides (PathVQA), and surgical videos (MedVideo), while performing multiple task types such as visual question answering, anatomical localization, modality classification, and medical report generation. This versatility highlights Fleming-VL's ability to serve as a unified foundation model for medical visual understanding.

\section{Conclusion}

We present Fleming-VL, a unified multimodal large language model that addresses the challenge of heterogeneous medical data understanding across 2D images, 3D volumetric scans, and temporal video sequences. Through systematic data curation targeting underrepresented modalities, scaled-up pretraining with long-context natural and medical data, and advanced post-training techniques combining supervised fine-tuning and group relative policy optimization, Fleming-VL achieves state-of-the-art performance across diverse medical understanding tasks. By publicly releasing model weights, training data, and evaluation protocols, we aim to promote transparent and reproducible progress in medical AI, contributing to safer AI deployment in high-stakes clinical environments and advancing toward unified vision-language models capable of supporting comprehensive clinical workflows.

\bibliographystyle{unsrt}  
\bibliography{references}  

\begin{thebibliography}{10}

\bibitem{liu2023visual}
Haotian Liu, Chunyuan Li, Qingyang Wu, and Yong~Jae Lee.
\newblock Visual instruction tuning.
\newblock {\em Advances in neural information processing systems}, 36:34892--34916, 2023.

\bibitem{liu2024improved}
Haotian Liu, Chunyuan Li, Yuheng Li, and Yong~Jae Lee.
\newblock Improved baselines with visual instruction tuning.
\newblock In {\em Proceedings of the IEEE/CVF conference on computer vision and pattern recognition}, pages 26296--26306, 2024.

\bibitem{yao2024minicpm}
Yuan Yao, Tianyu Yu, Ao~Zhang, Chongyi Wang, Junbo Cui, Hongji Zhu, Tianchi Cai, Haoyu Li, Weilin Zhao, Zhihui He, et~al.
\newblock Minicpm-v: A gpt-4v level mllm on your phone.
\newblock {\em arXiv preprint arXiv:2408.01800}, 2024.

\bibitem{dai2023instructblip}
Wenliang Dai, Junnan Li, Dongxu Li, Anthony Tiong, Junqi Zhao, Weisheng Wang, Boyang Li, Pascale~N Fung, and Steven Hoi.
\newblock Instructblip: Towards general-purpose vision-language models with instruction tuning.
\newblock {\em Advances in neural information processing systems}, 36:49250--49267, 2023.

\bibitem{wang2024qwen2}
Peng Wang, Shuai Bai, Sinan Tan, Shijie Wang, Zhihao Fan, Jinze Bai, Keqin Chen, Xuejing Liu, Jialin Wang, Wenbin Ge, et~al.
\newblock Qwen2-vl: Enhancing vision-language model's perception of the world at any resolution.
\newblock {\em arXiv preprint arXiv:2409.12191}, 2024.

\bibitem{li2023llava}
Chunyuan Li, Cliff Wong, Sheng Zhang, Naoto Usuyama, Haotian Liu, Jianwei Yang, Tristan Naumann, Hoifung Poon, and Jianfeng Gao.
\newblock Llava-med: Training a large language-and-vision assistant for biomedicine in one day.
\newblock {\em Advances in Neural Information Processing Systems}, 36:28541--28564, 2023.

\bibitem{li2024llava}
Jiajie Li, Garrett Skinner, Gene Yang, Brian~R Quaranto, Steven~D Schwaitzberg, Peter~CW Kim, and Jinjun Xiong.
\newblock Llava-surg: towards multimodal surgical assistant via structured surgical video learning.
\newblock {\em arXiv preprint arXiv:2408.07981}, 2024.

\bibitem{xu2025lingshu}
Weiwen Xu, Hou~Pong Chan, Long Li, Mahani Aljunied, Ruifeng Yuan, Jianyu Wang, Chenghao Xiao, Guizhen Chen, Chaoqun Liu, Zhaodonghui Li, et~al.
\newblock Lingshu: A generalist foundation model for unified multimodal medical understanding and reasoning.
\newblock {\em arXiv preprint arXiv:2506.07044}, 2025.

\bibitem{medgemini2024}
{Google DeepMind}.
\newblock Med-gemini: A family of multimodal medical ai models.
\newblock \url{https://research.google/blog/advancing-medical-ai-with-med-gemini/}, 2024.
\newblock Accessed: 2025-10-14.

\bibitem{liu2025fleming}
Chi Liu, Derek Li, Yan Shu, Robin Chen, Derek Duan, Teng Fang, and Bryan Dai.
\newblock Fleming-r1: Toward expert-level medical reasoning via reinforcement learning.
\newblock {\em arXiv preprint arXiv:2509.15279}, 2025.

\bibitem{wang2023huatuo}
Haochun Wang, Chi Liu, Nuwa Xi, Zewen Qiang, Sendong Zhao, Bing Qin, and Ting Liu.
\newblock Huatuo: Tuning llama model with chinese medical knowledge, 2023.

\bibitem{bai2024m3d}
Fan Bai, Yuxin Du, Tiejun Huang, Max Q-H Meng, and Bo~Zhao.
\newblock M3d: Advancing 3d medical image analysis with multi-modal large language models.
\newblock {\em arXiv preprint arXiv:2404.00578}, 2024.

\bibitem{chen20243d}
Hao Chen, Wei Zhao, Yingli Li, Tianyang Zhong, Yisong Wang, Youlan Shang, Lei Guo, Junwei Han, Tianming Liu, Jun Liu, et~al.
\newblock 3d-ct-gpt: Generating 3d radiology reports through integration of large vision-language models.
\newblock {\em arXiv preprint arXiv:2409.19330}, 2024.

\bibitem{chen2025surgllm}
Zhen Chen, Xingjian Luo, Kun Yuan, Jinlin Wu, Danny Chan, Nassir Navab, Hongbin Liu, Zhen Lei, and Jiebo Luo.
\newblock Surgllm: A versatile large multimodal model with spatial focus and temporal awareness for surgical video understanding.
\newblock {\em arXiv preprint arXiv:2509.00357}, 2025.

\bibitem{chen2024huatuogpt}
Junying Chen, Zhenyang Cai, Ke~Ji, Xidong Wang, Wanlong Liu, Rongsheng Wang, and Benyou Wang.
\newblock Towards medical complex reasoning with {LLM}s through medical verifiable problems.
\newblock In Wanxiang Che, Joyce Nabende, Ekaterina Shutova, and Mohammad~Taher Pilehvar, editors, {\em Findings of the Association for Computational Linguistics: ACL 2025}, pages 14552--14573, Vienna, Austria, July 2025. Association for Computational Linguistics.

\bibitem{flemingr1}
Chi Liu, Derek Li, Yan Shu, Robin Chen, Derek Duan, Teng Fang, and Bryan Dai.
\newblock Fleming-r1: Toward expert-level medical reasoning via reinforcement learning, 2025.

\bibitem{li2023chatdoctor}
Yunxiang Li, Zihan Li, Kai Zhang, Ruilong Dan, Steve Jiang, and You Zhang.
\newblock Chatdoctor: A medical chat model fine-tuned on a large language model meta-ai (llama) using medical domain knowledge.
\newblock {\em Cureus}, 15(6), 2023.

\bibitem{venigalla2022pubmed}
A~Venigalla, J~Frankle, and M~Carbin.
\newblock Pubmed gpt: A domain-specific large language model for biomedical text, 2022.

\bibitem{zhu2024mmedpo}
Kangyu Zhu, Peng Xia, Yun Li, Hongtu Zhu, Sheng Wang, and Huaxiu Yao.
\newblock Mmedpo: Aligning medical vision-language models with clinical-aware multimodal preference optimization.
\newblock {\em arXiv preprint arXiv:2412.06141}, 2024.

\bibitem{nguyen2024logra}
Duy~MH Nguyen, Nghiem~T Diep, Trung~Q Nguyen, Hoang-Bao Le, Tai Nguyen, Tien Nguyen, TrungTin Nguyen, Nhat Ho, Pengtao Xie, Roger Wattenhofer, et~al.
\newblock Logra-med: Long context multi-graph alignment for medical vision-language model.
\newblock {\em arXiv preprint arXiv:2410.02615}, 2024.

\bibitem{wu2024medical}
Junde Wu, Jiayuan Zhu, Yunli Qi, Jingkun Chen, Min Xu, Filippo Menolascina, and Vicente Grau.
\newblock Medical graph rag: Towards safe medical large language model via graph retrieval-augmented generation.
\newblock {\em arXiv preprint arXiv:2408.04187}, 2024.

\bibitem{jin2024matching}
Qiao Jin, Zifeng Wang, Charalampos~S Floudas, Fangyuan Chen, Changlin Gong, Dara Bracken-Clarke, Elisabetta Xue, Yifan Yang, Jimeng Sun, and Zhiyong Lu.
\newblock Matching patients to clinical trials with large language models.
\newblock {\em Nature communications}, 15(1):9074, 2024.

\bibitem{singhal2025toward}
Karan Singhal, Tao Tu, Juraj Gottweis, Rory Sayres, Ellery Wulczyn, Mohamed Amin, Le~Hou, Kevin Clark, Stephen~R Pfohl, Heather Cole-Lewis, et~al.
\newblock Toward expert-level medical question answering with large language models.
\newblock {\em Nature Medicine}, 31(3):943--950, 2025.

\bibitem{pan2025medvlm}
Jiazhen Pan, Che Liu, Junde Wu, Fenglin Liu, Jiayuan Zhu, Hongwei~Bran Li, Chen Chen, Cheng Ouyang, and Daniel Rueckert.
\newblock Medvlm-r1: Incentivizing medical reasoning capability of vision-language models (vlms) via reinforcement learning.
\newblock In {\em International Conference on Medical Image Computing and Computer-Assisted Intervention}, pages 337--347. Springer, 2025.

\bibitem{jiang2025omniv}
Songtao Jiang, Yuan Wang, Sibo Song, Yan Zhang, Zijie Meng, Bohan Lei, Jian Wu, Jimeng Sun, and Zuozhu Liu.
\newblock Omniv-med: Scaling medical vision-language model for universal visual understanding.
\newblock {\em arXiv preprint arXiv:2504.14692}, 2025.

\bibitem{xin2025med3dvlm}
Yu~Xin, Gorkem~Can Ates, Kuang Gong, and Wei Shao.
\newblock Med3dvlm: An efficient vision-language model for 3d medical image analysis.
\newblock {\em arXiv preprint arXiv:2503.20047}, 2025.

\bibitem{wang2025surgvidlm}
Guankun Wang, Wenjin Mo, Junyi Wang, Long Bai, Kun Yuan, Ming Hu, Jinlin Wu, Junjun He, Yiming Huang, Nicolas Padoy, et~al.
\newblock Surgvidlm: Towards multi-grained surgical video understanding with large language model.
\newblock {\em arXiv preprint arXiv:2506.17873}, 2025.

\bibitem{zhu2025internvl3}
Jinguo Zhu, Weiyun Wang, Zhe Chen, Zhaoyang Liu, Shenglong Ye, Lixin Gu, Hao Tian, Yuchen Duan, Weijie Su, Jie Shao, et~al.
\newblock Internvl3: Exploring advanced training and test-time recipes for open-source multimodal models.
\newblock {\em arXiv preprint arXiv:2504.10479}, 2025.

\bibitem{zhu2023multimodal}
Wanrong Zhu, Jack Hessel, Anas Awadalla, Samir~Yitzhak Gadre, Jesse Dodge, Alex Fang, Youngjae Yu, Ludwig Schmidt, William~Yang Wang, and Yejin Choi.
\newblock Multimodal c4: An open, billion-scale corpus of images interleaved with text.
\newblock {\em Advances in Neural Information Processing Systems}, 36:8958--8974, 2023.

\bibitem{siragusa2024medpix}
Irene Siragusa, Salvatore Contino, Massimo La~Ciura, Rosario Alicata, and Roberto Pirrone.
\newblock Medpix 2.0: a comprehensive multimodal biomedical dataset for advanced ai applications.
\newblock {\em arXiv preprint arXiv:2407.02994}, page~16, 2024.

\bibitem{luo2024fairclip}
Yan Luo, Min Shi, Muhammad~Osama Khan, Muhammad~Muneeb Afzal, Hao Huang, Shuaihang Yuan, Yu~Tian, Luo Song, Ava Kouhana, Tobias Elze, et~al.
\newblock Fairclip: Harnessing fairness in vision-language learning.
\newblock In {\em Proceedings of the IEEE/CVF Conference on Computer Vision and Pattern Recognition}, pages 12289--12301, 2024.

\bibitem{seyfioglu2024quilt}
Mehmet~Saygin Seyfioglu, Wisdom~O Ikezogwo, Fatemeh Ghezloo, Ranjay Krishna, and Linda Shapiro.
\newblock Quilt-llava: Visual instruction tuning by extracting localized narratives from open-source histopathology videos.
\newblock In {\em Proceedings of the IEEE/CVF Conference on Computer Vision and Pattern Recognition}, pages 13183--13192, 2024.

\bibitem{subramanian2020medicat}
Sanjay Subramanian, Lucy~Lu Wang, Sachin Mehta, Ben Bogin, Madeleine Van~Zuylen, Sravanthi Parasa, Sameer Singh, Matt Gardner, and Hannaneh Hajishirzi.
\newblock Medicat: A dataset of medical images, captions, and textual references.
\newblock {\em arXiv preprint arXiv:2010.06000}, 2020.

\bibitem{he2020pathvqa}
Xuehai He, Yichen Zhang, Luntian Mou, Eric Xing, and Pengtao Xie.
\newblock Pathvqa: 30000+ questions for medical visual question answering.
\newblock {\em arXiv preprint arXiv:2003.10286}, 2020.

\bibitem{zhang2023pmc}
Xiaoman Zhang, Chaoyi Wu, Ziheng Zhao, Weixiong Lin, Ya~Zhang, Yanfeng Wang, and Weidi Xie.
\newblock Pmc-vqa: Visual instruction tuning for medical visual question answering.
\newblock {\em arXiv preprint arXiv:2305.10415}, 2023.

\bibitem{liu2021slake}
Bo~Liu, Li-Ming Zhan, Li~Xu, Lin Ma, Yan Yang, and Xiao-Ming Wu.
\newblock Slake: A semantically-labeled knowledge-enhanced dataset for medical visual question answering.
\newblock In {\em 2021 IEEE 18th international symposium on biomedical imaging (ISBI)}, pages 1650--1654. IEEE, 2021.

\bibitem{ben2019vqa}
Asma Ben~Abacha, Sadid~A Hasan, Vivek~V Datla, Dina Demner-Fushman, and Henning M{\"u}ller.
\newblock Vqa-med: Overview of the medical visual question answering task at imageclef 2019.
\newblock In {\em Proceedings of CLEF (Conference and Labs of the Evaluation Forum) 2019 Working Notes}. 9-12 September 2019, 2019.

\bibitem{bae2024mimic}
Seongsu Bae, Daeun Kyung, Jaehee Ryu, Eunbyeol Cho, Gyubok Lee, Sunjun Kweon, Jungwoo Oh, Lei Ji, Eric Chang, Tackeun Kim, et~al.
\newblock Mimic-ext-mimic-cxr-vqa: A complex, diverse, and large-scale visual question answering dataset for chest x-ray images, 2024.

\bibitem{lau2018dataset}
Jason~J Lau, Soumya Gayen, Asma Ben~Abacha, and Dina Demner-Fushman.
\newblock A dataset of clinically generated visual questions and answers about radiology images.
\newblock {\em Scientific data}, 5(1):1--10, 2018.

\bibitem{johnson2019mimic}
Alistair~EW Johnson, Tom~J Pollard, Seth~J Berkowitz, Nathaniel~R Greenbaum, Matthew~P Lungren, Chih-ying Deng, Roger~G Mark, and Steven Horng.
\newblock Mimic-cxr, a de-identified publicly available database of chest radiographs with free-text reports.
\newblock {\em Scientific data}, 6(1):317, 2019.

\bibitem{demner2015preparing}
Dina Demner-Fushman, Marc~D Kohli, Marc~B Rosenman, Sonya~E Shooshan, Laritza Rodriguez, Sameer Antani, George~R Thoma, and Clement~J McDonald.
\newblock Preparing a collection of radiology examinations for distribution and retrieval.
\newblock {\em Journal of the American Medical Informatics Association}, 23(2):304--310, 2015.

\bibitem{guo2025deepseek}
Daya Guo, Dejian Yang, Haowei Zhang, Junxiao Song, Ruoyu Zhang, Runxin Xu, Qihao Zhu, Shirong Ma, Peiyi Wang, Xiao Bi, et~al.
\newblock Deepseek-r1: Incentivizing reasoning capability in llms via reinforcement learning.
\newblock {\em arXiv preprint arXiv:2501.12948}, 2025.

\bibitem{shao2024deepseekmath}
Zhihong Shao, Peiyi Wang, Qihao Zhu, Runxin Xu, Junxiao Song, Xiao Bi, Haowei Zhang, Mingchuan Zhang, YK~Li, Yang Wu, et~al.
\newblock Deepseekmath: Pushing the limits of mathematical reasoning in open language models.
\newblock {\em arXiv preprint arXiv:2402.03300}, 2024.

\bibitem{wang2025medgen}
Rongsheng Wang, Junying Chen, Ke~Ji, Zhenyang Cai, Shunian Chen, Yunjin Yang, and Benyou Wang.
\newblock Medgen: Unlocking medical video generation by scaling granularly-annotated medical videos.
\newblock {\em arXiv preprint arXiv:2507.05675}, 2025.

\bibitem{gornale2020comprehensive}
Shivanand~S Gornale, Pooja~U Patravali, and Prakash~S Hiremath.
\newblock A comprehensive digital knee x-ray image dataset for the assessment of osteoarthritis.
\newblock {\em JSM Biomed Imaging Data Pap}, 6:1012, 2020.

\bibitem{msoud_nickparvar_2021}
Msoud Nickparvar.
\newblock Brain tumor mri dataset, 2021.

\bibitem{vitale2020improving}
Santiago Vitale, Jos{\'e}~Ignacio Orlando, Emmanuel Iarussi, and Ignacio Larrabide.
\newblock Improving realism in patient-specific abdominal ultrasound simulation using {CycleGANs}.
\newblock {\em International Journal of Computer Assisted Radiology and Surgery}, 15(2):183--192, 2020.

\bibitem{ALDHABYANI2020104863}
Walid Al-Dhabyani, Mohammed Gomaa, Hussien Khaled, and Aly Fahmy.
\newblock Dataset of breast ultrasound images.
\newblock {\em Data in Brief}, 28:104863, 2020.

\bibitem{xu2022annotated}
Yiming Xu, Bowen Zheng, Xiaohong Liu, Tao Wu, Jinxiu Ju, Shijie Wang, Yufan Lian, Hongjun Zhang, Tong Liang, Ye~Sang, Rui Jiang, Guangyu Wang, Jie Ren, and Ting Chen.
\newblock Annotated ultrasound liver images, 2022.
\newblock Data set.

\bibitem{pacheco2020pad}
Andre~GC Pacheco, Gustavo~R Lima, Amanda~S Salomao, Breno Krohling, Igor~P Biral, Gabriel~G De~Angelo, F{\'a}bio~CR Alves~Jr, Jos{\'e}~GM Esgario, Alana~C Simora, Pedro~BC Castro, et~al.
\newblock Pad-ufes-20: A skin lesion dataset composed of patient data and clinical images collected from smartphones.
\newblock {\em Data in brief}, 32:106221, 2020.

\bibitem{cuadros2009eyepacs}
Jorge Cuadros and George Bresnick.
\newblock Eyepacs: an adaptable telemedicine system for diabetic retinopathy screening.
\newblock {\em Journal of diabetes science and technology}, 3(3):509--516, 2009.

\bibitem{ruckert2024rocov2}
Johannes R{\"u}ckert, Louise Bloch, Raphael Br{\"u}ngel, Ahmad Idrissi-Yaghir, Henning Sch{\"a}fer, Cynthia~S Schmidt, Sven Koitka, Obioma Pelka, Asma~Ben Abacha, Alba G.~Seco~de Herrera, et~al.
\newblock Rocov2: Radiology objects in context version 2, an updated multimodal image dataset.
\newblock {\em Scientific Data}, 11(1):688, 2024.

\bibitem{hu2024omnimedvqa}
Yutao Hu, Tianbin Li, Quanfeng Lu, Wenqi Shao, Junjun He, Yu~Qiao, and Ping Luo.
\newblock Omnimedvqa: A new large-scale comprehensive evaluation benchmark for medical lvlm.
\newblock In {\em Proceedings of the IEEE/CVF Conference on Computer Vision and Pattern Recognition}, pages 22170--22183, 2024.

\end{thebibliography}

\setcounter{section}{0}
\setcounter{figure}{0}    
\setcounter{table}{0}

\renewcommand{\thetable}{\Alph{table}}
\renewcommand{\thefigure}{\Alph{figure}}
\renewcommand{\thesection}{\Alph{section}}

\end{document}